\documentclass{article}

\usepackage{arxiv}

\usepackage[utf8]{inputenc} 
\usepackage[T1]{fontenc}    
\usepackage{hyperref}       
\usepackage{url}            
\usepackage{booktabs, multirow}       
\usepackage{amsfonts}       
\usepackage{nicefrac}       
\usepackage{microtype}      

\usepackage{amsmath, amssymb, amsthm, bm}
\usepackage{mathrsfs}
\usepackage{subcaption}
\usepackage[shortlabels]{enumitem}
\usepackage{float, graphicx, placeins}
\graphicspath{{figures/}}
\usepackage{algorithm}
\usepackage[noend]{algpseudocode}
\usepackage{multirow}
\usepackage{multicol}
\usepackage{tikz}

\theoremstyle{plain}
    \newtheorem{theorem}{Theorem}
    \newtheorem{proposition}{Proposition}

\theoremstyle{remark}

\theoremstyle{definition}
    \newtheorem{definition}{Definition}
    
\newcommand{\norm}[1]{\left\lVert#1\right\rVert}

\newcommand{\parenth}[1]{\left( #1 \right)}
\newcommand{\R}{\mathbb{R}}

\newcommand{\E}[1]{\mathbb{E}\left[ #1 \right]}

\title{Optimal Explanations of Linear Models}

\author{
    Dimitris Bersimas \\
    Sloan School of Management\\
    Massachusetts Institute of Technology\\
    Cambridge, MA 02139 \\
    \texttt{dbertsim@mit.edu} \\
    \And
    Arthur Delarue \\
    Operations Research Center\\
    Massachusetts Institute of Technology\\
    Cambridge, MA 02139 \\
    \texttt{adelarue@mit.edu} \\
    \And
    Patrick Jaillet \\
    Dep. of Electrical Engineering and Computer Science\\
    Massachusetts Institute of Technology\\
    Cambridge, MA 02139 \\
    \texttt{jaillet@mit.edu} \\
    \And
    Sebastien Martin \\
    Operations Research Center\\
    Massachusetts Institute of Technology\\
    Cambridge, MA 02139 \\
    \texttt{92sebastien@gmail.com} \\
}

\begin{document}

\maketitle

\begin{abstract}
  When predictive models are used to support complex and important decisions, the ability to explain a model's reasoning can increase trust, expose hidden biases, and reduce vulnerability to adversarial attacks. However, attempts at interpreting models are often ad hoc and application-specific, and the concept of interpretability itself is not well-defined. We propose a general optimization framework to create explanations for linear models. Our methodology decomposes a linear model into a sequence of models of increasing complexity using coordinate updates on the coefficients. Computing this decomposition optimally is a difficult optimization problem for which we propose exact algorithms and scalable heuristics. By solving this problem, we can derive a parametrized family of interpretability metrics for linear models that generalizes typical proxies, and study the tradeoff between interpretability and predictive accuracy.
\end{abstract}

\section{Introduction}

As machine learning models influence a growing fraction of everyday life, from cancer diagnoses to parole decisions to loan applications \cite{Mullainathan2017,Kleinberg2017,Berk2017}, individuals often want to understand the reasons for the decisions that affect them \cite{Dietvorst2016}.
Model interpretability is of significant interest to the machine learning community \cite{Freitas2014}, even though the lack of a well-defined concept of interpretability \cite{Lipton2016} means researchers often focus on proxies (e.g. sparsity or coefficient integrality in linear models).

Building ``simple'' models by optimizing interpretability proxies is a helpful but incomplete way to enhance interpretability.
In some cases, practitioners prefer more complex models (e.g. deeper decision trees) because they can be explained and justified in a more compelling way \cite{lavravc1999selected}.
Recent EU legislation \cite{Goodman2016} guarantees citizens a ``right to explanation,'' not a right to be affected only by sparse models.
But explaining a model in simple terms is a major challenge in interpretable machine learning, because it is typically an ad hoc, audience-specific process.
Formalizing the process of model explanation can yield ways to create models that are easier to explain, and rigorously quantify the tradeoff between interpretability and accuracy.


In this paper, we focus on linear models, and explore ways to decompose them into a sequence of interpretable coordinate updates.
We propose a general optimization framework to measure and optimize the interpretability of these sequences.
We then discuss how to create linear models with better explanations, leading to a natural set of interpretability metrics, and show that we can generalize various aspects of linear model interpretability. In particular,

\begin{itemize}
    \item Section~\ref{sec:paths} introduces \emph{coordinate paths} and motivates their use to explain linear models. 
    \item Section~\ref{sec:interpretabilityloss} presents a set of metrics to evaluate the interpretability of coordinate paths and extends them into interpretability metrics for models. This allows us to study the \emph{price of interpretability}, i.e., the Pareto front between accuracy and interpretability. We show that our metrics are consistent with existing approaches and exhibit desirable properties.
    \item Section~\ref{sec:computing} presents both optimal and scalable algorithms to compute coordinate paths and interpretable models.
    \item Section~\ref{sec:extensions} discusses various practical uses of our framework and other extensions.
\end{itemize}

\subsection{Related work}

Many interpretable machine learning approaches involve optimizing some characteristics of the model as proxies for interpretability. Examples include sparsity for linear models \cite{Hastie2015}, number of splits for decision trees \cite{Breiman1984}, number of subspace features for case-based reasoning \cite{Kim2015}, or depth for rule lists \cite{Letham2015,Yang2016a}. Some approaches optimize these proxies directly, while others fit auxiliary simple models to more complex black-box models \cite{Ribeiro2016,Friedman2001,Datta2016,Bastani2017,Lakkaraju2017,Bucila2006}.

In the specific case of linear models, the typical interpretability proxy of sparsity (small number of nonzero coefficients) has been a topic of extensive study over the past twenty years \cite{Hastie2015}. Sparse regression models can be trained using heuristics such as LASSO \cite{Tibshirani1996}, stagewise regression \cite{Taylor2015} or least-angle regression \cite{Efron2004}, or using scalable mixed-integer approaches \cite{Bertsimas2016}. More recently, another factor of interpretability in linear models has involved imposing integrality on the coefficients \cite{Jung2017,Ustun2016}, which allows to think of the output as tallying up points from each feature into a final score.

Training low-complexity models often affects predictive accuracy, and the tradeoff between the two can be difficult to quantify \cite{Breiman2001}. Similarly, the limitations of an ex post explanation relative to the original black box model can be difficult to explain to users \cite{Gilpin2018}. And it is not clear that practitioners always find models that optimize these proxies more interpretable \cite{lavravc1999selected}.
Recent landmark works \cite{Doshi-Velez2017,Lipton2016,Gilpin2018} have argued that any study of interpretability must include input from human users. 
The framework we propose is both human-driven and mathematically rigorous, as users can define their own understanding of interpretability and quantify the resulting tradeoff with accuracy.

\section{A Sequential View of Model Construction} \label{sec:paths}

Given a dataset with feature matrix ${X}\in\mathbb{R}^{n\times d}$ and labels ${y}\in\mathbb{R}^n$, a linear model is a vector of coefficients $\beta\in\mathbb{R}^d$, associated with a cost $c(\beta)$ that measures how well it fits the data, such as the mean-squared error $c(\beta)=(1/n)\|X\beta-y\|^2$ (potentially augmented with a regularization term for out-of-sample error).

We will motivate our approach to explaining linear models with a toy example. The goal is to predict a child's age \(y_\text{A}\) given height \(X_H\) and weight \(X_W\). The normalized features \(X_H\) and \(X_W\) have correlation \(\rho = 0.9\) and are both positively correlated with the normalized target. Solving the ordinary least squares problem yields optimal coefficients $\beta^*=(2.12, -0.94)$:
\begin{equation}\label{eq:model-linearexample}
    y_\text{A} = 2.12 \cdot X_H - 0.94 \cdot X_W + \varepsilon,
\end{equation}
with \(\varepsilon\) the error term. The mean squared error (MSE) of \(\beta^*\) is \(c(\beta^*) =\E{\varepsilon^2} =0.25\).

\subsection{Coordinate paths}

We propose a framework to construct an explanation of $\beta^*$ by decomposing the model into a sequence of interpretable building blocks.
In particular, we consider sequences of of linear models leading to $\beta^*$ where each model is obtained by changing one coefficient from the preceding model.
We choose these \emph{coordinate steps} because they correspond to the natural idea of adding a feature or updating an existing coefficient, and we will show they have interesting properties. We discuss other potential steps in Section~\ref{sec:extensions}.
Table~\ref{tab:three-decompositions} shows 3 possible decompositions of $\beta^*$ into coordinate steps.
This is a natural way to decompose $\beta^*$, notice for example that decomposition \ref{tab:decomposition1} corresponds to introducing the model coefficient by coefficient.

\begin{table}[h!]
    \centering
    \caption{Three decompositions of $\beta^*$ into a sequence of coordinate steps.}
    \label{tab:three-decompositions}
    \begin{subtable}[c]{0.33\linewidth}
    \centering
    \begin{tabular}{lr}
    \toprule
    $\beta=(\beta_H, \beta_W)$ & $c(\beta)$\\
    \midrule
    $\beta_0=(0,0)$ & $2.04$\\
    $\beta_1=(2.12,0)$ & $1.13$\\
    $\beta_2=(2.12,-0.94)$ & $0.25$\\
    \bottomrule
    \end{tabular}
    \caption{Two steps, starting with $X_H$}
    \label{tab:decomposition1}
    \end{subtable}\hfill%
    \begin{subtable}[c]{0.33\linewidth}
    \centering
    \begin{tabular}{lr}
    \toprule
    $\beta=(\beta_H, \beta_W)$ & $c(\beta)$\\
    \midrule
    $\beta'_0=(0,0)$ & $2.04$\\
    $\beta'_1=(0,-0.94)$ & $4.74$\\
    $\beta'_2=(2.12,-0.94)$ & $0.25$\\
    \bottomrule
    \end{tabular}
    \caption{Two steps, starting with $X_W$}
    \label{tab:decomposition2}
    \end{subtable}\hfill%
    \begin{subtable}[c]{0.33\linewidth}
    \centering
    \begin{tabular}{lr}
    \toprule
    $\beta=(\beta_H, \beta_W)$ & $c(\beta)$\\
    \midrule
    $\hat{\beta}_0=(0,0)$ & $2.04$\\
    $\hat{\beta}_1=(1.70,0)$ & $0.60$\\
    $\hat{\beta}_2=(1.70,-0.94)$ & $0.43$\\
    $\hat{\beta}_3=(2.12,-0.94)$ & $0.25$\\
    \bottomrule
    \end{tabular}
    \caption{Three steps, changing $X_H$ twice}
    \label{tab:decomposition3}
    \end{subtable}
    \end{table}

We refer to these sequences of models as \emph{coordinate paths}. Formally, we define a coordinate path of length $K$ as a sequence of $K$ models $\bm{\beta} = (\beta_1, \cdots, \beta_K)$ such that $\beta_k \in \mathcal{S}(\beta_{k-1})$ for all $1\le k \le K$, where $\beta_0 = 0$, and $\mathcal{S}(\beta)$ is the set of linear models that are one coordinate step away from $\beta$, i.e., $\mathcal{S}(\beta)=\{\theta\in\mathbb{R}^d : \norm{\beta-\theta}_0\le 1\}$. $\mathcal{P}_K$ is the set of all coordinate paths of length $K$, and $\mathcal{P} = \cup_{K=1}^\infty \mathcal{P}_K$ is the set of all finite coordinate paths.
An \emph{explanation} of a model $\beta$ is a coordinate path $\bm{\beta}\in \mathcal{P}$ such that the last model is $\beta$. $\mathcal{P}_K(\beta)$ is the set of explanations of $\beta$ of length $K$ (potentially empty), and $\mathcal{P}(\beta)$ is the set of all possible explanations of $\beta$ (typically infinite).

The examples in Table~\ref{tab:three-decompositions} are all explanations of $\beta^*$, and it is natural to ask which is the most useful or interpretable one. Formally speaking, it is of interest to define an \emph{interpretability loss} $\mathcal{L}(\cdot)$ on the space of coordinate paths $\mathcal{P}$, such that $\mathcal{L}(\bm{\beta})<\mathcal{L}(\bm{\beta'})$ when $\bm{\beta}$ is more interpretable than $\bm{\beta'}$.
Then finding the best possible explanation for any model $\beta$ can be written as the optimization problem
\[
    \min_{\bm{\beta} \in \mathcal{P}(\beta)} \mathcal{L}(\bm{\beta}).
\]
For any path interpretability loss $\mathcal{L}(\cdot)$, it is then easy to consider the interpretability loss $\mathcal{L}(\beta)$ of a \emph{model} $\beta$ as the interpretability loss of the best explanation $\bm{\beta}\in\mathcal{P}(\beta)$, i.e.
\begin{equation}
    \mathcal{L}(\beta) = \begin{cases}
                \infty, &\text{if } \mathcal{P}(\beta) = \emptyset,\\
                \min_{\bm{\beta} \in \mathcal{P}(\beta)} \mathcal{L}(\bm{\beta}), &\text{otherwise.}
            \end{cases}
    \label{eq:path-to-model-interpretability}
\end{equation}

How to select a path interpretability loss $\mathcal{L}(\bm{\beta})$? A natural choice is to consider that an explanation is better if it is shorter. Formally, we define the \emph{path complexity} loss $\mathcal{L}_c(\bm{\beta}) = |\bm{\beta}|$, corresponding to the length of the coordinate path. For any model $\beta$, we can define the corresponding interpretability loss
\[
\mathcal{L}_c(\beta)=\min_{\bm{\beta} \in \mathcal{P}(\beta)} \mathcal{L}_c(\bm{\beta})=\min_{\bm{\beta} \in \mathcal{P}(\beta)}|\bm{\beta}|,
\]
which we call \emph{model complexity} (minimum number of coordinate steps required to reach $\beta$). Interestingly, for any model $\beta$, $\mathcal{L}_c(\beta)$ corresponds to the number of non-zero coefficients of $\beta$. The natural metric of coordinate path length thus recovers the usual interpretability proxy of model sparsity.

Consider the different coordinate paths in Table~\ref{tab:three-decompositions}. If we use the interpretability loss $\mathcal{L}_c$, i.e., if we consider shorter paths (and thus sparser models) to be more interpretable, then \(\bm{\beta}\) (path~\ref{tab:decomposition1}) and \(\bm{\beta}'\) (path~\ref{tab:decomposition2}) are equally interpretable. Though both paths verify $c(\beta_2)=c(\beta'_2)=0.25$, we notice that $c(\beta_1)=1.13<c(\beta'_1)=4.74$. Indeed, \(\beta'_1\) is a particularly inaccurate model, as weight is actually positively correlated with age. Since a coordinate path represents an explanation of the final model, the costs of intermediate models should play a role in quantifying the interpretability of a path; higher costs should be penalized. The path complexity loss $\mathcal{L}_c(\cdot)$ does not consider intermediate model costs at all, and therefore cannot capture this effect.

\subsection{Incrementality}

To explore alternatives to path complexity, consider the example of greedy coordinate paths. where the next model at each step is chosen by minimizing the cost $c(\cdot)$:
\begin{equation} \label{eq:model-greedy}
    \beta^{\mathrm{greedy}}_{k+1} \in \arg\min \left\{ c(\beta),\,  \beta \in \mathcal{S}(\beta^{\mathrm{greedy}}_k) \right\} \quad \forall k \geq 1.
\end{equation}
This approach is appealing from an explanation standpoint, because we always select the coordinate step which most improves the model. However, many steps may be required to obtain an accurate model (slow convergence). Returning to toy example ~\eqref{eq:model-linearexample} and considering only paths of length 2, we compute the greedy coordinate path $\bm{\beta}^\mathrm{greedy}$ by solving~\eqref{eq:model-greedy} twice \cite{Taylor2015}. Comparing to $\bm{\beta}$ from Table~\ref{tab:decomposition1}, we have $c(\beta^{\mathrm{greedy}}_1) = 0.42 < 1.13=c(\beta_1)$, but $c(\beta^{\mathrm{greedy}}_2) = 0.39 > 0.25=c(\beta_2)$. The improvement of the first model comes at the expense of the second step.

\begin{figure}[h]
    \centering
    \includegraphics[width=0.4\columnwidth]{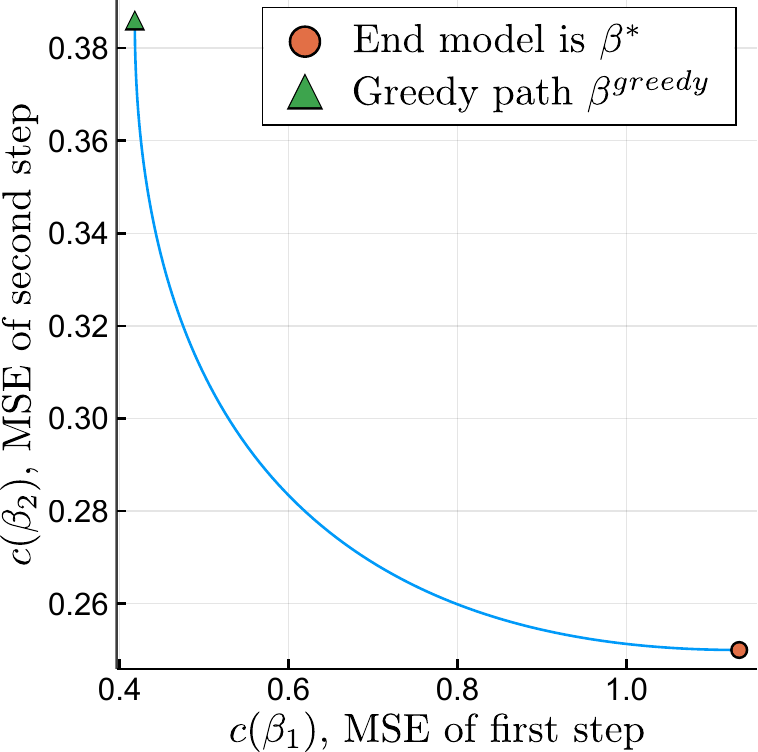}
    \caption{Tradeoff between the cost of the first and second models for a coordinate path of length 2 on problem~\eqref{eq:model-linearexample}.}
    \label{fig:models-2stepsgammatradeoff}
\end{figure}

Deciding which of the two paths $\bm{\beta}$ and $\bm{\beta}^\mathrm{greedy}$ is more interpretable is a hard question.
It highlights the tradeoff between the desirable incrementality of the greedy approach and the optimality of the second model. For paths of length 2, there is a continuum of solutions that trade off MSE in the first and second steps, shown in Figure~\ref{fig:models-2stepsgammatradeoff}. The next section introduces interpretability losses that formalize this tradeoff.

\section{Defining an Interpretability Loss}\label{sec:interpretabilityloss}

\subsection{Coherent Interpretability Losses}

In example~\eqref{eq:model-linearexample}, comparing the interpretability of $\bm{\beta}$ and $\bm{\beta'}$ is easy, because they have the same length and $\bm{\beta}$ has a better cost than $\bm{\beta'}$ at each step. In contrast, comparing the interpretability of $\bm{\beta}$ and $\bm{\beta}^\mathrm{greedy}$ is not trivial, because $\bm{\beta}$ has a better final cost, but $\bm{\beta}^\mathrm{greedy}$ has a better initial cost.

We can define the \emph{cost sequence} of a coordinate path $\bm{\beta} \in \mathcal{P}_K$ as the infinite sequence $(c_1, c_2, \cdots)$ such that $c_k=c(\beta_k)$ if $k\leq K$, and $c_k=0$ otherwise. Then we call a path interpretability loss $\mathcal{L}(\cdot)$ \emph{coherent} if the following conditions hold for any two paths $\bm{\beta}, \bm{\beta'}\in\mathcal{P}$ with cost sequences $\bm{c}$ and $\bm{c'}$.
\begin{enumerate}[(a)]
\item If $\bm{c}=\bm{c'}$, then $\mathcal{L}(\bm{\beta})=\mathcal{L}(\bm{\beta'})$\label{def:coherent-costdependent}.
\item If $c_k\le c'_k\;\forall k$, then $\mathcal{L}(\bm{\beta})\le\mathcal{L}(\bm{\beta'})$\label{def:coherent-pareto}.
\end{enumerate}

Condition~\ref{def:coherent-costdependent} means that in our modeling framework, the interpretability of a path depends only on the sequence of costs along that path. Condition~\ref{def:coherent-pareto} formalizes the intuition that paths with fewer steps or better steps are more interpretable. Under any coherent interpretability loss $\mathcal{L}(\cdot)$ in toy example~\eqref{eq:model-linearexample}, $\bm{\beta}$ is more interpretable than $\bm{\beta'}$, but $\bm{\beta}$ may be more or less interpretable than $\bm{\beta}^{\text{greedy}}$ depending on the specific choice of coherent interpretability loss.

In addition, consider a path $\bm{\beta}\in\mathcal{P}_K$ and remove its last step to obtain a new path $\bm{\beta'}\in\mathcal{P}_{K-1}$. This is equivalent to setting the $K$-th element of the cost sequence $\bm{c}(\bm{\beta})$ to zero. Since $c(\cdot)\ge 0$, we have that $\bm{c}(\bm{\beta})\le\bm{c}(\bm{\beta})$, which implies $\mathcal{L}(\bm{\beta'})\le\mathcal{L}(\bm{\beta})$. In other words, under a coherent interpretability loss, removing a step from a coordinate path can only make the path more interpretable. We also notice that the path complexity $\mathcal{L}_c$ (sparsity) is a coherent path interpretability loss. 

\subsection{A Coherent Model Interpretability Loss}

Condition~\ref{def:coherent-pareto} states that a path with at least as good a cost at each step as another path must be at least as interpretable. This notion of Pareto dominance suggests a natural path interpretability loss:
\[
\mathcal{L}_{\alpha}(\bm{\beta})=\sum_{k=1}^\infty\alpha_k \bm{c}(\bm{\beta})_k = \sum_{k=1}^{|\bm{\beta}|}\alpha_k c(\beta_k).
\]
In other words, the interpretability loss $\mathcal{L}_{\alpha}$ of a path $\bm{\beta}$ is the weighted sum of the costs of all steps in the path. This loss function is trivially coherent and extremely general. It is specified by the infinite sequence of parameters $\bm{\alpha}$, which specify the relative importance of the accuracy of each step in the model for the particular application at hand.

Defining a family of interpretability losses with infinitely many parameters allows for significant modeling flexibility, but it is also cumbersome and overly general. We therefore propose to select $\alpha_k=\gamma^k$ for all $k$, replacing the infinite sequence of parameters $(\alpha_1, \alpha_2, \ldots)$ with a single parameter $\gamma>0$. In this case, following \ref{eq:path-to-model-interpretability}, we propose the following interpretability loss function on the space of models.

\begin{definition}[Model interpretability]\label{def:model-interpretability}
    Given a model $\beta\in\mathbb{R}^d$, its interpretability loss $\mathcal{L}_{\gamma}(\beta)$ is given by
    \begin{equation}
        \mathcal{L}_{\gamma}(\beta) = \begin{cases}
            \infty, &\text{if } \mathcal{P}(\beta) = \emptyset,\\
            \min\limits_{\bm{\beta} \in \mathcal{P}(\beta)} \mathcal{L}_{\gamma}(\bm{\beta})=\sum\limits_{k=1}^{|\bm{\beta}|}\gamma^kc(\beta_k), &\text{otherwise.}
        \end{cases}
    \end{equation}
\end{definition}

By definition, $\mathcal{L}_{\gamma}$ is a coherent interpretability loss. The parameter $\gamma$ captures the tradeoff between favoring more incremental models or models with a low complexity, as formalized in Theorem~\ref{thm:model-gammainfinite}.

\begin{theorem}[Consistency of interpretability measure]\label{thm:model-gammainfinite}
    Assume that $c(\cdot)$ is bounded and nonnegative.
    \begin{enumerate}[(a)]
        \item \label{enum:model-thm3} Let $\beta^+,~\beta^- \in \mathbb{R}^d$ with $\mathcal{L}_{\mathrm{complexity}}(\beta^+) < \mathcal{L}_{\mathrm{complexity}}(\beta^-)$, or $\mathcal{L}_{\mathrm{complexity}}(\beta^+) = \mathcal{L}_{\mathrm{complexity}}(\beta^-)$ and $c(\beta^+) < c(\beta^-)$.
        \begin{equation}
            \lim_{\gamma \rightarrow \infty} \mathcal{L}_{\bm{\gamma}}(\beta^-) - \mathcal{L}_{\bm{\gamma}}(\beta^+) = +\infty.
        \end{equation}
        \item \label{enum:model-thm4} 
        Given models $\beta^+, \beta^- \in \mathbb{R}^d$, if there is $\bm{\beta^+} \in \mathcal{P}(\beta^+)$ such that $\bm{c}(\bm{\beta^+}) \preceq \bm{c}(\bm{\beta^-})$ for all $\bm{\beta^-} \in \mathcal{P}(\beta^-)$, then
        \begin{equation}
            \lim_{\gamma \rightarrow 0} \mathcal{L}_{\bm{\gamma}}(\beta^-) - \mathcal{L}_{\bm{\gamma}}(\beta^+) \geq 0.
        \end{equation}
    \end{enumerate}
\end{theorem}

Intuitively, in the limit $\gamma \rightarrow +\infty$, \ref{enum:model-thm3} states that the most interpretable models are the ones with minimal complexity, or minimal costs if their complexity is the same. \ref{enum:model-thm4} states that in the limit $\gamma \rightarrow 0$ the most interpretable models are the ones that can be constructed with greedy steps. All proofs are provided in the supplement.

\subsection{The Price of Interpretability}\label{sec:priceofinterpretability}

Given the metric of interpretability defined above, we want to compute models that are Pareto-optimal with respect to $c(\cdot)$ and $\mathcal{L}_{\gamma}(\cdot)$ (more generally $\mathcal{L}_{\alpha}(\cdot)$). Computing these models can be challenging, as our definition of model interpretability requires to optimize over paths of any length. We can get around this if we can at least find the most interpretable path of a fixed length $K$, i.e.,

\begin{equation}\label{eq:fixedpathopt}
    \min_{\bm{\beta} \in \mathcal{P}_K} \mathcal{L}_\alpha(\bm{\beta}) = \sum_{k=1}^K \alpha_k c(\beta_k)
\end{equation}

Indeed, the following result shows that we can compute Pareto-optimal solutions by solving a sequence of optimization problems \eqref{eq:fixedpathopt} for various $K$.

\begin{proposition}[Price of interpretability]\label{prop:model-priceofinterpretability}
    Pareto-optimal models that minimize the interpretability loss $\mathcal{L}_{\alpha}$ and the cost $c(\cdot)$ can be computed by solving the following optimization problem:
    \begin{equation} \label{eq:minimize-over-k}
        \min_{K \geq 0} \left( \min_{\bm{\beta} \in \mathcal{P}_K} c(\beta_K) + \lambda \sum_{k=1}^K \alpha_k c(\beta_k) \right) ,
    \end{equation}
    where $\lambda \in \mathcal{R}$ is a tradeoff parameter between cost and interpretability. 
\end{proposition}

\begin{figure}[h]
    \centering
    \includegraphics[width=0.7\columnwidth]{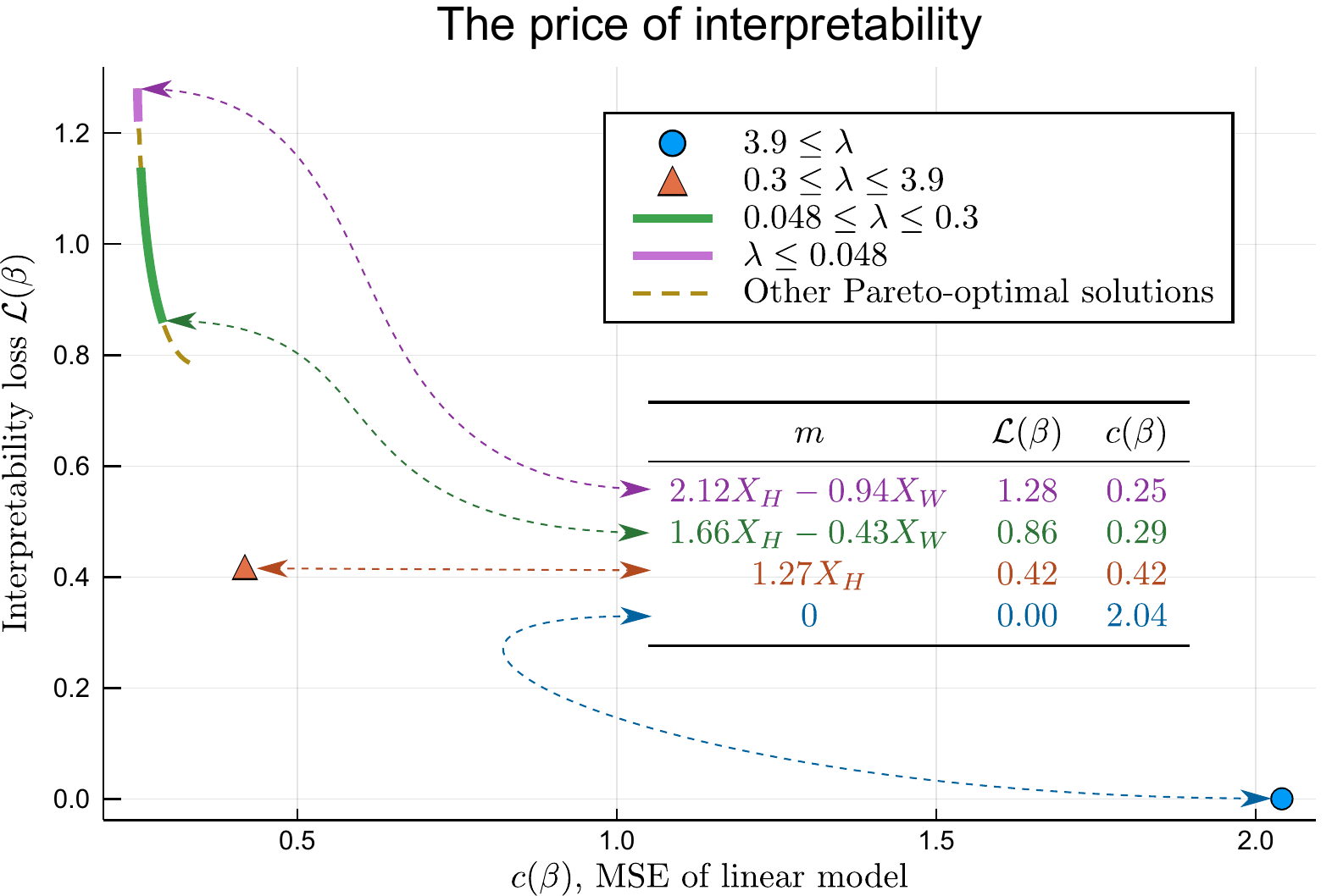}
    \caption{Pareto front between interpretability loss $\mathcal{L}(\beta)=\mathcal{L}_{\gamma}(\beta)$ (with $\gamma=1$) and cost $c(\beta)$ on the toy OLS problem \eqref{eq:model-linearexample}, computed by varying $\lambda$ in \eqref{eq:minimize-over-k}. The dashed line represents Pareto-optimal solutions that cannot be computed by this weighted-sum method \cite{Kim2005}. Note that the front is discontinuous. The number of steps in the corresponding optimal coordinate paths is respectively 0, 1, 2, 3 for the blue, orange, green and purple segments. The inset table describes several interesting Pareto-optimal models.}
    \label{fig:models-priceofinterpretability}
\end{figure}

Notice that the inner minimization problem in~\eqref{eq:minimize-over-k} is simply problem~\eqref{eq:fixedpathopt} with appropriate modifications of the coefficients $(\alpha_1, \ldots, \alpha_K)$. We can use this decomposition to compute the price of interpretability in the toy problem~\eqref{eq:model-linearexample}, with the interpretability loss $\mathcal{L}_{\gamma}$ chosen such that $\gamma=1$. Figure~\ref{fig:models-priceofinterpretability} shows all Pareto-optimal models with respect to cost (MSE) and interpretability loss.

By defining the general framework of coordinate paths and a natural family of coherent interpretability loss functions, we can understand exactly how much we gain or lose in terms of accuracy when we choose a more or less interpretable model. Our framework thus provides a principled way to answer a central question of the growing literature on interpretability in machine learning.


\section{Computing the Price of Interpretability}\label{sec:computing}

\subsection{Algorithms}

\paragraph{Optimal.} Given the step function $\mathcal{S}(\cdot)$ and the convex quadratic cost function $c(\cdot)$, problem~\eqref{eq:fixedpathopt} can be written as a convex integer optimization problem using special ordered sets of type 1 (SOS-1 constraints), and solved using Gurobi or CPLEX for small problems:
\begin{equation}\label{eq:linreg}
\min\sum_{k=1}^Kc(\beta_{k}) \quad \text{s.t.}\quad\text{SOS-1}(\beta_{k+1}-\beta_{k}) \;\forall\, 0\le k < K,
\end{equation}
where $\beta_0$ designates the starting linear model.

\paragraph{Local improvement.} In higher-dimensional settings, or when $K$ grows large, the formulation above may no longer scale. Thus it is of interest to develop a fast heuristic for such instances.

A feasible solution $\bm{\beta}$ to problem~\eqref{eq:linreg} can be written as a vector of indices $i\in\{1,\ldots,d\}^K$ and a vector of values $\delta\in\mathbb{R}^K$, such that for $0\le k < K$,
\[
\beta_{k+1}^i = \begin{cases}\beta^i_{k} + \delta_k, & \text{if }i = i_k\\
\beta_{k}^i, & \text{if }i\neq i_k.
\end{cases}
\]

The vector of indices $i$ encodes which coefficients are modified at each step, while the vector of values $\delta$ encodes the value of each modified coefficient. Thus problem \eqref{eq:linreg} can be rewritten as
\begin{equation}
\min_{i} \min_{\delta}C\left(i, \delta\right):=\sum_{k=1}^K c\left(\beta_0 + \sum_{j=1}^k\delta_je_{i_j}\right),
\end{equation}
where $e_i$ designates the $i$-th unit vector. The inner minimization problem is an ``easy'' convex quadratic optimization problem, while the outer minimization problem is a ``hard'' combinatorial optimization problem. We propose the following local improvement heuristic for the outer problem: given a current vector of indices $i$, we randomly sample one step $\kappa$ in the coordinate path. Keeping all $i_k$ constant for $k\neq\kappa$, we iterate through all $d$ possible values of $i_{\kappa}$ and obtain $d$ candidate vectors $\hat{i}$. For each candidate, we solve the inner minimization problem and keep the one with lowest cost. A general version of this algorithm, where we sample not one but $q$ steps at each iteration, is provided in the supplement.

\subsection{Results}

\paragraph{Optimal vs heuristic.}

In order to empirically evaluate the local improvement heuristic, we run it with different batch sizes $q$ on a small real dataset, with 100 rows and 6 features (after one-hot encoding of categorical features). The goal is to predict the perceived prestige (from a survey) of a job occupation given features about it, including education level, salary, etc.

Given this dataset, we first compute the optimal coordinate path of length $K=10$. We then test our local improvement heuristic on the same dataset. Given the small size of the problem, in the complete formulation a provable global optimum is found by Gurobi in about 5 seconds. To be useful, we would like our local improvement heuristic to find a good solution significantly faster. We show convergence results of the heuristic for different values of the batch size parameter $q$ in Table~\ref{tab:convergence}. 
For both batch sizes, the local improvement heuristic converges two orders of magnitude faster than Gurobi. With a batch size $q=2$, the solution found is optimal.

\begin{table}
\centering
\caption{Convergence of local improvement heuristics for different batch sizes $q$.}
\label{tab:convergence}
\begin{tabular}{lrr}
\toprule
Method & Time to convergence (s) & Optimality gap (\%)\\
\midrule
Exact & $5.078$ & $0.00$\\
Local improvement ($q=1$) & $0.004$ & $0.02$\\
Local improvement ($q=2$) & $0.019$ & $0.00$\\
\bottomrule
\end{tabular}
\end{table}

\paragraph{Insights from a real dataset.} We now explore the results of the presented approach on a dataset from the 1998-1999 California test score dataset. Each data point represents a school, and the variable of interest is the average standardized test score of students from that school. The ten continuous features and the target variables are centered and rescaled to have unit variance.

In our example, we assume that we already have a regression model available to predict the number of trips: it was trained using only the percentage of students qualifying for a reduced-price lunch. This model has an MSE of 0.122 (compared to an optimal MSE of 0.095). We would like to update this model in an interpretable way given the availability of all features in the dataset. In our framework, this corresponds to problem~\eqref{eq:linreg} where $\beta_0$ is no longer 0 but the available starting model.

\begin{figure}
\begin{subfigure}[c]{0.45\columnwidth}
\centering
\begingroup
\footnotesize
\begin{tabular}{lccccr}
\toprule
& \multicolumn{4}{c}{Feature}\\
 & MealPct & AvgInc & ELPct & ExpnStu & MSE\\
\midrule
$\beta_0$ & $-0.87$ & - & - & - & 0.122\\
$\beta_1$ & \vrule & ${0.23}$ & - & - & 0.122\\
$\beta_2$ & ${-0.59}$ & \vrule & - & - & 0.117\\
$\beta_3$ & \vrule & \vrule & ${-0.18}$ & - & 0.099\\
$\beta_4$ & \vrule & \vrule & \vrule & ${0.07}$ & 0.097\\
\bottomrule
\end{tabular}
\endgroup
\caption{Coordinate path from old model to new model. MealPct is the percentage of students qualifying for reduced-price lunch, AvgInc is the average income, ELPct is the percentage of English Learners, ExpnStu is the expenditure per student.}
\label{fig:path-caschool-details}
\end{subfigure}\hfill
\begin{subfigure}[c]{0.5\columnwidth}
\includegraphics[width=\columnwidth]{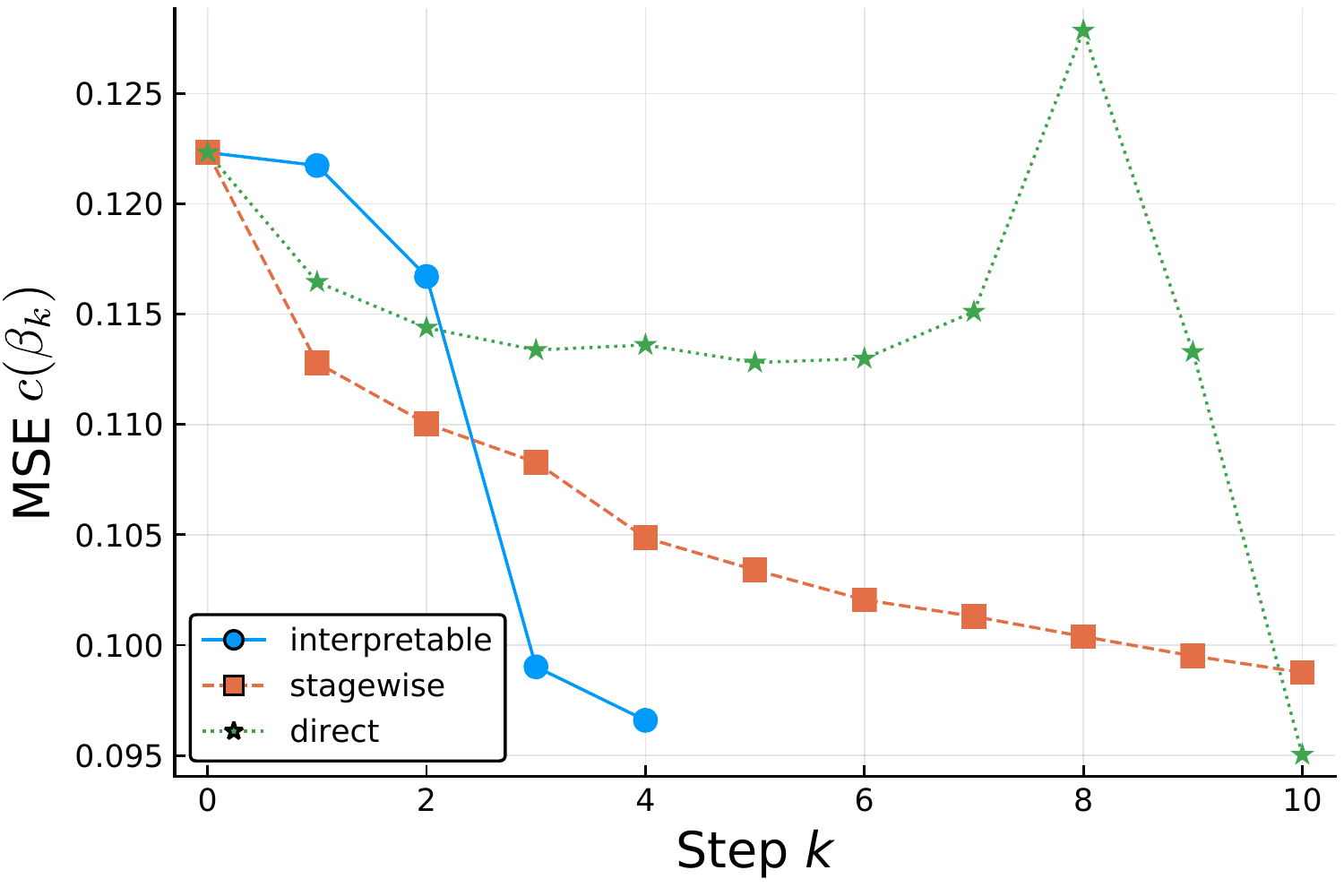}
\caption{Comparison between coordinate path and other approaches}
\label{fig:path-caschool-comparison}
\end{subfigure}
\caption{Example of a Pareto-efficient coordinate path. On the left we see the benefits of each coefficient modification. On the right we compare the coordinate path with the forward stagewise path which greedily selects the best $\beta_{k+1}$ given $\beta_k$, and the ``direct'' path, which adds the optimal least squares coefficients one by one. The direct method is only good when all the coefficients have been added, whereas the greedy approach is good at first but then does not converge. The coordinate path is willing to make some suboptimal steps in preparation for very cost-improving steps.}
\label{fig:path-caschool}
\end{figure}

In Figure~\ref{fig:path-caschool} we study one particular coordinate path on the Pareto front of interpretability and efficiency. The path (and the new model it leads to) is shown in Figure~\ref{fig:path-caschool-details}. It can be obtained from the original model in just four steps. First we add the district average income with a positive coefficient, then we correct the coefficient for reduced-price lunch students to account for this new feature, and finally we add the percentage of English learners and the per-student spending. The final model has an MSE of 0.097 which is near-optimal. When we compare this path to other methods (see Figure~\ref{fig:path-caschool-comparison}) we see that our interpretable formulation allows us to find a good tradeoff between a greedy formulation and a formulation that just sets the coefficients to their final values one by one.

\section{A General Framework}\label{sec:extensions}

\subsection{Different Steps for Different Notions of Interpretability}

So far, we have focused exclusively on paths in which linear models are constructed in a series of coordinate steps. However, the choice of what constitutes a step is ultimately a modeling choice which encodes what a user in a particular application considers a simple building block. Choosing a different step function $\mathcal{S}(\cdot)$ can lead to other notions of interpretability. For instance, choosing $\mathcal{S}_\mathrm{SLIM}(\beta)=\{\theta: \norm{\beta-\theta}_0\le1, \norm{\beta-\theta}_1\in\mathbb{Z}\}$ imposes integer coordinate updates at each step. This is related to the notion of interpretability introduced by the score-based methods of Ustun and Rudin \cite{Ustun2016}. Another way to think about score-based methods is to choose $\mathcal{S}'_\mathrm{SLIM}(\beta)=\{\theta: \norm{\beta-\theta}_0\le1, \norm{\beta-\theta}_1\in\{0,1\}\}$, which imposes that each step adds one point to the scoring system. The fundamental idea of optimally decomposing models into a sequence of simple building blocks is general, and can be applied not only to more general linear models (e.g. ridge regression or logistic regression by suitably modifying the cost $c$) but also to other machine learning models in general (for example, a decision tree can be decomposed into a sequence of successive splits).

\subsection{Human-in-the-loop Model Selection}
Viewing a coordinate path as a nested sequence of models of increasing complexity can be useful in the context of human-in-the-loop analytics. Consider the problem of selecting a linear model by a human decision maker. For example, consider a city planner that would like to understand bike-sharing usage in Porto, by training a linear model using a dataset from the UCI ML repository \cite{Fanaee-T2014}, where each of the 731 data points represents a particular day (18 features about weather, time of year, etc.), and the variable of interest is the number of trips recorded by the bike-sharing system on that day. The decision-maker may prefer a sparse model, but may not know the exact desired level of sparsity. Given a discrete distribution on the choice of the level of sparsity $K$ : $p_k = \mathbb{P}(\{ k \text{ will be chosen by the decision maker}\})$, we can choose $\alpha_k = p_k$ and solve \eqref{eq:fixedpathopt} to find paths $\bm{\beta}$ that minimize the expected cost $\mathbb{E}_k[c(\beta_k)]$.

In Table~\ref{tab:interpretable-path}, we show the path $\bm{\beta}$ obtained by assuming that the desired level of sparsity is uniformly random between 1 and 7. We can compare the result to a sequence of linear models obtained using LASSO to select an increasing set of features, shown in Table~\ref{tab:bike-lasso}. The expected costs are respectively 0.145 and 0.150, and the MSE is essentially the same as each step. The two sequences also use almost the same features, in a similar order. However, the coordinate path can be read much more easily: because only one coefficient changes at each step, the whole path can be described with $\Theta(K)$ parameters, while the path constructed using LASSO/sparse regression needs $\Theta(K^2)$ parameters.

\begin{figure}
\begin{subfigure}[t]{0.5\linewidth}
\centering
\scriptsize
\begin{tabular}{lcccccccr}
\toprule
&  &  &  & Season: &  & Weekday: & Weather: & \\
& Ftemp & Day & Hum & Summer & Wind & Sunday & Clear & MSE\\
\midrule
$\beta_1$ & 0.56 & & & & & & & 0.293\\
$\beta_2$ & \vrule & 0.55 & & & & & & 0.146\\
$\beta_3$ & \vrule & \vrule & -0.20 & & & & & 0.129\\
$\beta_4$ & \vrule & \vrule & \vrule & 0.15 & & & & 0.119\\
$\beta_5$ & \vrule & \vrule & \vrule & \vrule & -0.14 & & & 0.110\\
$\beta_6$ & \vrule & \vrule & \vrule & \vrule & \vrule & -0.07 & & 0.108\\
$\beta_7$ & \vrule & \vrule & \vrule & \vrule & \vrule & \vrule & 0.06 & 0.107\\
\bottomrule
\end{tabular}
\caption{Interpretable path for the bike-sharing dataset.}
\label{tab:interpretable-path}
\end{subfigure}
\begin{subfigure}[t]{0.5\linewidth}
\centering
\scriptsize
\begin{tabular}{lcccccccr}
\toprule
&  &  & Season: & Weather: &  & Season: &  & \\
& Ftemp & Day & Spring & Clear & Hum & Summer & Wind & MSE\\
\midrule
$\beta_1$ & 0.65 & & & & & & & 0.289\\
$\beta_2$ & 0.55 & 0.54 & & & & & & 0.146\\
$\beta_3$ & 0.50 & 0.52 & -0.10 & & & & & 0.143\\
$\beta_4$ & 0.47 & 0.52 & -0.12 & 0.18 & & & & 0.128\\
$\beta_5$ & 0.49 & 0.51 & -0.13 & 0.09 & -0.15 & & & 0.121\\
$\beta_6$ & 0.50 & 0.55 & -0.07 & 0.10 & -0.14 & 0.13 & & 0.114\\
$\beta_7$ & 0.49 & 0.54 & -0.06 & 0.08 & -0.18 & 0.14 & -0.14 & 0.107\\
\bottomrule
\end{tabular}
\caption{Sequence of models with an increasing number of features obtained via LASSO.}
\label{tab:bike-lasso}
\end{subfigure}
\caption{Comparison of interpretable path and sequence of models of increasing sparsity selected by LASSO. Ftemp is the ``feels like'' temperature, Day is the number of days since data collection began, Hum is humidity, Wind is wind speed. Season, Weather and Weekday are categorical variables.}
\end{figure}

\FloatBarrier

\section*{Acknowledgements}
Research funded in part by ONR grant N00014-18-1-2122.

\small
\bibliographystyle{unsrt}
\bibliography{references}

\begin{thebibliography}{10}

\bibitem{Mullainathan2017}
Sendhil Mullainathan and Ziad Obermeyer.
\newblock {Does machine learning automate moral hazard and error?}
\newblock {\em American Economic Review}, 107(5):476--480, 2017.

\bibitem{Kleinberg2017}
Jon Kleinberg, Himabindu Lakkaraju, Jure Leskovec, Jens Ludwig, and Sendhil
  Mullainathan.
\newblock {Human decisions and machine predictions}.
\newblock {\em The quarterly journal of economics}, 133(1):237--293, 2017.

\bibitem{Berk2017}
Richard Berk.
\newblock {An impact assessment of machine learning risk forecasts on parole
  board decisions and recidivism}.
\newblock {\em Journal of Experimental Criminology}, 13(2):193--216, 2017.

\bibitem{Dietvorst2016}
Berkeley~J Dietvorst, Joseph~P Simmons, and Cade Massey.
\newblock {Overcoming algorithm aversion: People will use imperfect algorithms
  if they can (even slightly) modify them}.
\newblock {\em Management Science}, 64(3):1155--1170, nov 2016.

\bibitem{Freitas2014}
Alex~A. Freitas.
\newblock {Comprehensible classification models}.
\newblock {\em ACM SIGKDD Explorations Newsletter}, 15(1):1--10, 2014.

\bibitem{Lipton2016}
Zachary~C. Lipton.
\newblock {The Mythos of Model Interpretability}.
\newblock {\em arXiv preprint arXiv:1606.03490}, 2016.

\bibitem{lavravc1999selected}
Nada Lavra{\v{c}}.
\newblock Selected techniques for data mining in medicine.
\newblock {\em Artificial intelligence in medicine}, 16(1):3--23, 1999.

\bibitem{Goodman2016}
Bryce Goodman and Seth Flaxman.
\newblock {European Union regulations on algorithmic decision-making and a
  "right to explanation"}.
\newblock pages 1--9, 2016.

\bibitem{Hastie2015}
Trevor Hastie, Robert Tibshirani, and Martin Wainwright.
\newblock {\em {Statistical learning with sparsity: the lasso and
  generalizations}}.
\newblock CRC press, 2015.

\bibitem{Breiman1984}
Leo Breiman.
\newblock {\em {Classification and regression trees}}.
\newblock New York: Routledge, 1984.

\bibitem{Kim2015}
Been Kim, Cynthia Rudin, and Julie Shah.
\newblock {The Bayesian Case Model: A Generative Approach for Case-Based
  Reasoning and Prototype Classification}.
\newblock In {\em Neural Information Processing Systems (NIPS) 2014}, 2014.

\bibitem{Letham2015}
Benjamin Letham, Cynthia Rudin, Tyler~H. McCormick, and David Madigan.
\newblock {Interpretable classifiers using rules and bayesian analysis:
  Building a better stroke prediction model}.
\newblock {\em Annals of Applied Statistics}, 9(3):1350--1371, 2015.

\bibitem{Yang2016a}
Hongyu Yang, Cynthia Rudin, and Margo Seltzer.
\newblock {Scalable Bayesian Rule Lists}.
\newblock In {\em Proceedings of the 34th International Conference on Machine
  Learning}, 2017.

\bibitem{Ribeiro2016}
Marco~Tulio Ribeiro, Sameer Singh, and Carlos Guestrin.
\newblock {“Why Should I Trust You?” Explaining the Predictions of Any
  Classifier}.
\newblock In {\em Proceedings of the 22nd ACM SIGKDD international conference
  on knowledge discovery and data mining}, pages 1135--1144, 2016.

\bibitem{Friedman2001}
Jerome Friedman, Trevor Hastie, and Robert Tibshirani.
\newblock {\em {The elements of statistical learning}}.
\newblock Springer series in statistics New York, NY, USA:, 2001.

\bibitem{Datta2016}
Anupam Datta, Shayak Sen, and Yair Zick.
\newblock {Algorithmic Transparency via Quantitative Input Influence :}.
\newblock In {\em 2016 IEEE Symposium on Security and Privacy}, 2016.

\bibitem{Bastani2017}
Hamsa Bastani, Osbert Bastani, and Carolyn Kim.
\newblock {Interpreting Predictive Models for Human-in-the-Loop Analytics}.
\newblock {\em arXiv preprint arXiv:1705.08504}, pages 1--45, 2018.

\bibitem{Lakkaraju2017}
Himabindu Lakkaraju, Ece Kamar, Rich Caruana, and Jure Leskovec.
\newblock {Interpretable {\&} Explorable Approximations of Black Box Models}.
\newblock {\em FAT/ML}, jul 2017.

\bibitem{Bucila2006}
Cristian Bucilǎ, Rich Caruana, and Alexandru Niculescu-Mizil.
\newblock {Model compression}.
\newblock In {\em Proceedings of the 12th ACM SIGKDD international conference
  on Knowledge discovery and data mining - KDD '06}, page 535, New York, New
  York, USA, 2006. ACM, ACM Press.

\bibitem{Tibshirani1996}
Robert~J. Tibshirani.
\newblock {Regression shrinkage and selection via the lasso}.
\newblock {\em Journal of the Royal Statistical Society. Series B
  (Methodological)}, pages 267--288, 1996.

\bibitem{Taylor2015}
Jonathan Taylor and Robert~J. Tibshirani.
\newblock {Statistical learning and selective inference}.
\newblock {\em Proceedings of the National Academy of Sciences},
  112(25):7629--7634, jun 2015.

\bibitem{Efron2004}
Bradley Efron, Trevor Hastie, Iain Johnstone, and Robert Tibshirani.
\newblock {Least Angle Regression}.
\newblock {\em Annals of Statistics}, 32(2):407--499, apr 2004.

\bibitem{Bertsimas2016}
Dimitris Bertsimas, Angela King, and Rahul Mazumder.
\newblock {Best subset selection via a modern optimization lens}.
\newblock {\em Annals of Statistics}, 44(2):813--852, 2016.

\bibitem{Jung2017}
Jongbin Jung, Connor Concannon, Ravi Shroff, Sharad Goel, and Daniel~G
  Goldstein.
\newblock {Simple rules for complex decisions}.
\newblock feb 2017.

\bibitem{Ustun2016}
Berk Ustun and Cynthia Rudin.
\newblock {Supersparse linear integer models for optimized medical scoring
  systems}.
\newblock {\em Machine Learning}, 102(3):349--391, 2016.

\bibitem{Breiman2001}
Leo Breiman.
\newblock {Statistical modeling: The two cultures}.
\newblock {\em Statistical science}, 16(3):199--231, 2001.

\bibitem{Gilpin2018}
Leilani~H Gilpin, David Bau, Ben~Z Yuan, Ayesha Bajwa, Michael Specter, and
  Lalana Kagal.
\newblock {Explaining Explanations : An Approach to Evaluating Interpretability
  of Machine Learning}.
\newblock {\em arXiv preprint arXiv:1806.00069}, 2018.

\bibitem{Doshi-Velez2017}
Finale Doshi-Velez and Been Kim.
\newblock {Towards A Rigorous Science of Interpretable Machine Learning}.
\newblock {\em arXiv preprint arXiv:1702.08608}, (Ml):1--13, 2017.

\bibitem{Kim2005}
I.~Y. Kim and O.~L. {De Weck}.
\newblock {Adaptive weighted-sum method for bi-objective optimization: Pareto
  front generation}.
\newblock {\em Structural and Multidisciplinary Optimization}, 29(2):149--158,
  2005.

\bibitem{Fanaee-T2014}
Hadi Fanaee-T and Joao Gama.
\newblock {Event labeling combining ensemble detectors and background
  knowledge}.
\newblock {\em Progress in Artificial Intelligence}, 2(2-3):113--127, 2014.

\end{thebibliography}

\appendix 

\section{Proof of Theorem 1}

\begin{proof}[Proof of part (a)]
    As $c(\cdot)$ is bounded, we have $c_{\max} \in \R$ such that $0 < c(\cdot) \leq c_{\max}$.

    Let $\bm{m^+} \in \mathcal{P}(m^+)$ be a path of optimal length to the model $m^+$, i.e., $|\bm{m^+}| = \mathcal{L}_c(m^+)$. . Let $\bm{m^-} \in \mathcal{P}(m^-)$ be any path leading to $m^-$ (not necessarily of optimal length). By assumption, we have $|\bm{m^-}| \geq |\bm{m^+}|$, and by definition of model interpretability, we have $\mathcal{L}_\gamma(m^+) \leq \mathcal{L}_\gamma(\bm{m^+})$. Therefore we obtain:
    \begin{align}
        &\mathcal{L}_\gamma(\bm{m^-}) - \mathcal{L}_\gamma(m^+) \geq \mathcal{L}_\gamma(\bm{m^-}) - \mathcal{L}_\gamma(\bm{m^+}) \\
        &\quad = \sum_{k=1}^{|\bm{m^-}|} \gamma^k\, c(m^-_k) - \sum_{k=1}^{|\bm{m^+}|} \gamma^k\, c(m^+_k) \label{eq:model-proofthmconsistency-1}\\
        &\quad = \gamma^{|\bm{m^+}|} \parenth{\sum_{k=1}^{|\bm{m^+}| - 1} \frac{1}{\gamma^{|\bm{m^+}|-k}}\, \parenth{c(m^-_k) - c(m^+_k)} +  \parenth{c(m^-_{|\bm{m^+}|}) - c(m^+_{|\bm{m^+}|})} + \sum_{k=|\bm{m^+}| + 1}^{|\bm{m^-}|} \gamma^{k-|\bm{m^+}|}\, c(m^-_k) } \label{eq:model-proofthmconsistency-2}\\
        &\quad \geq \gamma^{|\bm{m^+}|} \parenth{- c_{\max} \sum_{k=1}^{|\bm{m^+}| - 1} \frac{1}{\gamma^{|\bm{m^+}|-k}} + \left( c(m^-_{|\bm{m^+}|}) - c(m^+) \right) +  \sum_{k=|\bm{m^+}| + 1}^{|\bm{m^-}|} \gamma^{k-|\bm{m^+}|}\, c(m^-_k)},\label{eq:model-proofthmconsistency-3}
    \end{align} 
    where \eqref{eq:model-proofthmconsistency-1} follows from the definition of model interpretability, \eqref{eq:model-proofthmconsistency-2} is just a development of the previous equation, and \eqref{eq:model-proofthmconsistency-3} just bounds the first sum and uses $m^+_{|\bm{m^+}|} = m^+$ for the middle term.

    If $\mathcal{L}_c(m^+) < \mathcal{L}_c(m^-)$, we have $|\bm{m^+}| < |\bm{m^-}|$, and therefore the last sum in \eqref{eq:model-proofthmconsistency-3} is not empty and for $\gamma \geq 1$ we can bound it:
    \begin{equation}
        \sum_{k=|\bm{m^+}| + 1}^{|\bm{m^-}|} \gamma^{k-|\bm{m^+}|}\, c(m^-_k) \geq \gamma^{|\bm{m^-}|-|\bm{m^+}|} \, c(m_{|\bm{m^-}|}^-) \geq \gamma \, c(m^-).
    \end{equation}

    Therefore, for $\gamma \geq 1$ we have:
    \begin{equation}
        \mathcal{L}_\gamma(\bm{m^-}) - \mathcal{L}_\gamma(m^+) \geq \gamma^{|\bm{m^+}|} \left( \gamma \, c(m^-) - c(m^+ ) - c_{max} \sum_{k=1}^{|\bm{m^+}| - 1} \frac{1}{\gamma^{|\bm{m^+}|-k}}\right).
    \end{equation}

    This bound is valid for all the path $\bm{m^-}$ leading to $m^-$, in particular the one with optimal interpretability loss, therefore we have (for $\gamma \geq 1$): 
    \begin{equation}
        \mathcal{L}_\gamma(m^-) - \mathcal{L}_\gamma(m^+) \geq \gamma^{|\bm{m^+}|} \left( \gamma \, c(m^-) - c(m^+ ) - c_{max} \sum_{k=1}^{|\bm{m^+}| - 1} \frac{1}{\gamma^{|\bm{m^+}|-k}}\right).
    \end{equation}
    which implies (as $c(m^-) > 0$):
    \begin{equation}
        \lim_{\gamma \rightarrow +\infty} \mathcal{L}_\gamma(m^-) - \mathcal{L}_\gamma(m^+) = +\infty
    \end{equation}

    We now look at the case  $\mathcal{L}_c(m^+) = \mathcal{L}_c(m^-)$ and $c(m^+) < c(m^-)$. For $\gamma \geq 1$, we can easily bound parts of equation \eqref{eq:model-proofthmconsistency-3}:
    \begin{equation}
        c(m^-_{|\bm{m^+}|}) + \sum_{k=|\bm{m^+}| + 1}^{|\bm{m^-}|} \gamma^{k-|\bm{m^+}|}\, c(m^-_k) \geq \gamma^{|\bm{m^-}|-|\bm{m^+}|} c(m^-_{|\bm{m^-}|})\geq c(m^-).
    \end{equation}

    Putting it back into \eqref{eq:model-proofthmconsistency-3}, we obtain (for $\gamma \geq 1$)
    \begin{equation}
        \mathcal{L}_\gamma(\bm{m^-}) - \mathcal{L}_\gamma(m^+) \geq \gamma^{|\bm{m^+}|} \left( \left( c(m^-) - c(m^+) \right) - c_{max} \sum_{k=1}^{|\bm{m^+}| - 1} \frac{1}{\gamma^{|\bm{m^+}|-k}}\right).
    \end{equation}

    This bound is independent of the path $\bm{m^-}$ leading to $m^-$, therefore we have 
    \begin{equation}
        \mathcal{L}_\gamma(m^-) - \mathcal{L}_\gamma(m^+)  \geq \gamma^{|\bm{m^+}|} \left( \left( c(m^-) - c(m^+) \right) - c_{max} \sum_{k=1}^{|\bm{m^+}| - 1} \frac{1}{\gamma^{|\bm{m^+}|-k}}\right) \rightarrow_{\gamma \rightarrow +\infty} +\infty,
    \end{equation}
    which ends the proof.
\end{proof}

\begin{proof}[Proof of part (b)]
    Consider two paths $\bm{m^+}, \bm{m^-} \in \mathcal{P}$, such that $\bm{c}(\bm{m^+}) \preceq \bm{c}(\bm{m^-})$. By definition of the lexicographic order, either the two paths are the same (in that case the theorem is trivial), or there exist $K \geq 1$ such that:
    \begin{equation*}
        \begin{cases}
            \bm{c}(\bm{m^+})_k = \bm{c}(\bm{m^-})_k \quad \forall k < K \\
            \bm{c}(\bm{m^+})_K < \bm{c}(\bm{m^-})_K .
        \end{cases}
    \end{equation*}
    
    We have:
    \begin{align}
        \mathcal{L}_\gamma(\bm{m^-}) - \mathcal{L}_\gamma(\bm{m^+}) &= \sum_{k=1}^{|\bm{m^-}|} \gamma^k\, c(m^-_k) - \sum_{k=1}^{|\bm{m^+}|} \gamma^k\, c(m^+_k) \\
        &= \sum_{k=1}^{\infty} \gamma^k\, \left( \bm{c}(\bm{m^-})_k - \bm{c}(\bm{m^+})_k \right) \label{eq:model-proofthmconsistency-4}\\
        &= \sum_{k=1}^{K-1} \gamma^k\, \left( \bm{c}(\bm{m^-})_k - \bm{c}(\bm{m^+})_k \right) + \gamma^K\left( \bm{c}(\bm{m^-})_K - \bm{c}(\bm{m^+})_K \right) \nonumber\\
        &\qquad \quad + \sum_{k=K+1}^{\infty} \gamma^k\, \left( \bm{c}(\bm{m^-})_k - \bm{c}(\bm{m^+})_k \right)\\
        &= \gamma^K \left( \bm{c}(\bm{m^-})_K - \bm{c}(\bm{m^+})_K + \sum_{k=K+1}^{\infty} \gamma^{k-K}\, \left( \bm{c}(\bm{m^-})_k - \bm{c}(\bm{m^+})_k \right)\right), \label{eq:model-proofthmconsistency-5}
    \end{align}
    where \eqref{eq:model-proofthmconsistency-4} just applies the definition of the sequence $\bm{c}$, and \eqref{eq:model-proofthmconsistency-5} uses $\bm{c}(\bm{m^+})_k = \bm{c}(\bm{m^-})_k \quad \forall k < K$.
    
    The term inside the parenthesis in \eqref{eq:model-proofthmconsistency-5} converges to $\bm{c}(\bm{m^-})_K - \bm{c}(\bm{m^+})_K > 0$ when $\gamma \rightarrow 0$, as the paths are finite. Therefore  
    \begin{equation}
        \lim_{\gamma \rightarrow 0} \mathcal{L}_\gamma(\bm{m^-}) - \mathcal{L}_\gamma(\bm{m^+}) \geq 0,
    \end{equation}
    which completes the proof. The very end of the theorem is an immediate consequence.
\end{proof}

\section{Proof of Proposition 1}

\begin{proof}
    First, a solution of 
    \begin{equation*}
        \min_{m \in \mathcal{M}} \left(c(m) + \lambda \mathcal{L}(m)\right)
    \end{equation*}
    is Pareto optimal between the cost $c(\cdot)$ and the interpretability $\mathcal{L}_{\alpha}(\cdot)$ as it corresponds to the minimization of a weighted sum of the objectives. Furthermore, we can write
    \begin{align*}
        \min_{m \in \mathcal{M}} \left(c(m) + \lambda \mathcal{L}_{\alpha}(m)\right)
        =& \min_{m \in \mathcal{M}} \left(c(m) + \lambda \min_{\bm{m}\in \mathcal{P}(m)} \mathcal{L}_{\alpha}(\bm{m})\right)
        = \min_{m \in \mathcal{M},\, \bm{m}\in \mathcal{P}(m)} \left(c(m) + \lambda \mathcal{L}_{\alpha}(\bm{m})\right)\\
        =& \min_{m \in \mathcal{M},\, K \geq 0,\, \bm{m}\in \mathcal{P}_K(m)}
        \left(c(m_K) + \lambda \sum_{k=1}^K\alpha_k c(m_k)\right)\\
        =& \min_{K \geq 0,\, \bm{m}\in \mathcal{P}_K} \left(c(m_K) + \lambda \sum_{k=1}^K\alpha_k c(m_k)\right).
\end{align*}
\end{proof}

\section{Local improvement algorithm}

\begin{algorithm}
\caption{Local improvement heuristic. Inputs: regression cost function $c(\cdot)$; starting vector of indices $i^0$. Parameters: $q\in\mathbb{N}$ controls the size of the batch, $T\in\mathbb{N}$ controls the number of iterations.}\label{algo:local-improvement}
\begin{algorithmic}[1]
\Function{LocalImprovement}{$c(\cdot)$, $\bm{i}^0$, $q$, $T$}
\For{$1 \le t \le T$}
\State $\bm{i}^* \gets \bm{i}^0$
\State $\bm{\delta}^* \gets \arg\min_{\bm{\delta}} C\left(\bm{i}^0, \bm{\delta}\right)$
\State $C^* \gets C\left(\bm{i}^0, \bm{\delta}^*\right)$
\State Randomly select $\mathcal{K}=\{\kappa_1, \ldots, \kappa_q\} \subset \{1,\ldots,K\}$ \Comment{subset of cardinality $q$}
\State $\bm{\hat{i}} \gets \bm{i}^*$
\State $\bm{\hat{\delta}} \gets \bm{\delta}^*$
\For{$(f_1, \ldots, f_q) \in \{1, \ldots, d\}^q$}
\For{$1 \le p \le q$}
\State $\hat{i}_{\kappa_p} = f_p$
\EndFor
\State $\bm{\hat{\delta}} \gets \arg\min_{\bm{\delta}} C\left(\bm{\hat{i}}, \bm{\delta}\right)$
\If{$C\left(\bm{\hat{i}}, \bm{\hat{\delta}}\right) < C^*$}
\State $C^* \gets C\left(\bm{\hat{i}}, \bm{\hat{\delta}}\right)$
\State $\bm{i}^* \gets \bm{\hat{i}}$
\State $\bm{\delta}^* \gets \hat{\delta}$
\EndIf
\EndFor
\EndFor
\State \textbf{return} $\bm{i}^*, \bm{\delta}^*$
\EndFunction
\end{algorithmic}
\end{algorithm}
\end{document}